\documentclass[lettersize,journal]{IEEEtran}
\usepackage{amsmath,amsfonts}
\usepackage{algorithmic}
\usepackage{algorithm}
\usepackage{array}
\usepackage[caption=false,font=normalsize,labelfont=sf,textfont=sf]{subfig}
\usepackage{textcomp}
\usepackage{stfloats}
\usepackage{url}
\usepackage{verbatim}
\usepackage{graphicx}
\usepackage{subfig}
\usepackage{cite}
\usepackage{booktabs}
\usepackage{xcolor}

\newcommand{\firstRevision}[1]{\textcolor{black}{#1}}
\newcommand{\secondRevision}[1]{\textcolor{black}{#1}}
\newcommand{\rebuttal}[1]{\textcolor{black}{#1}}

\begin{document}

\title{\firstRevision{Leveraging AI Agents for Autonomous Networks: A Reference Architecture and Empirical Studies}}

% \author{Binghan Wu, Shoufeng Wang, Ya-Qin Zhang, Joseph Sifakis, Ye Ouyang}
\author{
    \IEEEauthorblockN{Binghan Wu\textsuperscript{1}, Shoufeng Wang\textsuperscript{1}, Yunxin Liu\textsuperscript{2}, Ya-Qin Zhang\textsuperscript{2}, Joseph Sifakis\textsuperscript{3}, Ye Ouyang\textsuperscript{1}}
    
    \IEEEauthorblockA{\textsuperscript{1}\textit{AsiaInfo Technologies Limited}, Beijing, China}
    \IEEEauthorblockA{\textsuperscript{2}\textit{Institute for AI Industry Research (AIR), Tsinghua University}, Beijing, China}
    \IEEEauthorblockA{\textsuperscript{3}\textit{Verimag}, \textit{Université Grenoble Alpes}, Grenoble, France}
}

% The paper headers
\markboth{Journal of \LaTeX\ Class Files,~Vol.~14, No.~8, August~2021}%
{Shell \MakeLowercase{\textit{et al.}}: A Sample Article Using IEEEtran.cls for IEEE Journals}

\IEEEpubid{}
% Remember, if you use this you must call \IEEEpubidadjcol in the second
% column for its text to clear the IEEEpubid mark.

\maketitle

\begin{abstract}\color{black}
The evolution toward Level 4 (L4) Autonomous Networks (AN) represents a strategic inflection point in telecommunications, where networks must transcend reactive automation to achieve genuine cognitive capabilities—fulfilling AN's vision of self-configuring, self-healing, and self-optimizing systems that deliver zero-wait, zero-touch, and zero-fault services. This work bridges the gap between architectural theory and operational reality by implementing Joseph Sifakis's AN Agent reference architecture in a functional cognitive system, deploying coordinated proactive-reactive runtimes driven by hybrid knowledge representation. Through an empirical case study of a Radio Access Network (RAN) Link Adaptation (LA) Agent, we validate this framework's performance. Specifically, the system demonstrates sub-10 ms real-time control in 5G NR sub-6 GHz environments. Empirical results show a 4\% increase in downlink throughput over Outer Loop Link Adaptation (OLLA) algorithms for enhanced mobile broadband (eMBB). Furthermore, for the ultra-reliable low-latency communication (URLLC) scenario, the agent achieves an 85\% reduction in Block Error Rate (BLER). These improvements confirm the architecture's viability in overcoming traditional autonomy barriers and advancing critical L4-enabling capabilities toward next-generation objectives. 
\end{abstract}

\section{Introduction}
\IEEEPARstart{A}{utonomous} Networks (AN), a purpose-specific telecommunications technology pioneered by the TM Forum (TMF) in 2019, target networks with intrinsic \textit{self-configuration}, \textit{self-healing}, and \textit{self-optimization} capabilities—collectively termed the Three-Self Capabilities~\cite{chinadaily2025autonomous}. These fundamental properties enable the realization of \textit{zero-wait}, \textit{zero-touch}, and \textit{zero-fault} network services, known as the Three-Zero Objectives, which collectively deliver optimal user experiences while maximizing resource utilization throughout the entire network lifecycle. By strategically integrating emerging general-purpose technologies including artificial intelligence (AI), digital twins, and big data analytics, AN not only transforms conventional network operations but fundamentally reorients value creation paradigms from traditional device-centric and management-centric models toward customer-oriented, service-driven, and business-focused frameworks.

The TMF's standardized six-level maturity continuum (L0–L5) provides telecommunications operators with a structured methodology for assessing AN evolutionary progress, where ascending levels denote progressively enhanced automation and cognitive capabilities. Currently, the global telecommunications industry is undergoing a critical transition phase, advancing from mid-to-low autonomy stages (L2/L3: partial or conditional autonomy) toward the pivotal threshold of L4 (high-level autonomy). Major operators worldwide have established comprehensive roadmaps aligning with this evolutionary trajectory. Nevertheless, AN implementation confronts significant systemic challenges as a complex systems engineering endeavor. Persistent limitations in AI algorithmic complexity and technological maturity have engendered operational bottlenecks metaphorically described as ``intelligence plateau'' and ``acceleration resistance.'' Traditional machine learning-based approaches embed task-specific intelligence into AN workflows but exhibit inherent deficiencies in genuine autonomous decision-making. Concurrently, Large Language Model (LLM)-based AI copilots demonstrate enhanced knowledge provision and recommendation capabilities through human-AI collaboration frameworks, yet remain constrained by limited proactive intervention capacities. Fundamentally, both methodologies encounter autonomy ceilings at conditional levels (L3), necessitating the development of agents with advanced self-governance mechanisms to achieve L4 operational targets.

The conceptual foundation for autonomous agents traces back to Marvin Minsky's seminal 1986 work \textit{The Society of Mind}~\cite{1986Society}, which introduced the notion of computational entities exhibiting intelligent behavior. Contemporary autonomous agents represent sophisticated computational systems capable of demonstrating contextually appropriate behaviors within specific operational environments. These systems integrate core competencies spanning \textit{environmental perception}, \textit{cognitive reasoning}, \textit{strategic decision-making}, and \textit{physical actuation}, continuously enhancing performance through experiential learning and adaptive recalibration. By leveraging accumulated knowledge repositories, engaging in self-directed knowledge acquisition, and participating in collaborative multi-agent ecosystems, they autonomously accomplish predefined operational objectives.

While reinforcement learning (RL) agents previously dominated autonomous systems—demonstrating adaptive behaviors through iterative environmental interactions yet fundamentally limited by their reactive nature and lack of higher-order cognition—the emergence of generative AI has catalyzed a paradigm shift toward LLM-based agents, marking a significant evolutionary leap toward genuinely autonomous and cognitively sophisticated AI systems. Modern LLM-based agents exhibit \textit{proactive cognitive engagement}, requiring minimal supervisory intervention while integrating multifaceted capabilities including knowledge synthesis, associative memory, logical reasoning, multi-step planning, and risk-aware decision-making. These systems actively utilize digital tools and environmental interfaces for contextual judgment and execution, with continuous learning mechanisms enabling knowledge expansion, adaptation to novel operational scenarios, and pursuit of emergent objectives—collectively establishing the technical underpinnings for high-level operational autonomy.

This research contributes two principal advancements to the field: first, it presents a novel dual-driver AN Agent reference architecture that synthesizes external network environmental dynamics with internal network requirements, accompanied by the first documented implementation case study based on this architectural framework. Second, it provides empirical validation through a RAN LA Agent functioning at the network element level. This case study represents pioneering real-world demonstrations of AN Agent capabilities, quantitatively validating L4 autonomy through network performance enhancements.

\section{From AI Agent to AN Agent}
\subsection{Current AI Agent Architectures}

Contemporary AI agents achieve environmental interaction through world models that explicitly represent environmental dynamics as learnable transition functions or implicitly encode them within latent neural representations. LeCun's architecture~\cite{LeCun2022APT} exemplifies explicit approaches with modular predictive models for real-time control, while implicit paradigms like DayDreamer~\cite{wu2022daydreamerworldmodelsphysical} encode dynamics in latent structures for sample-efficient robot learning. The A$^3$T framework~\cite{yang2024reactmeetsactrelanguage} advances this paradigm through autonomous trajectory synthesis, achieving 96\% success in embodied tasks without manual intervention.

Modern reasoning agents increasingly rely on LLMs as cognitive cores. The hybrid LAW framework~\cite{hu2023languagemodelsagentmodels} integrates symbolic modules for logical consistency in precision tasks, while ReAct~\cite{2022ReAct} dynamically interleaves cognition with environmental interaction to mitigate hallucinations. Self-Consistency~\cite{wang2023selfconsistencyimproveschainthought} enhances reliability through diverse reasoning path sampling, boosting accuracy by +17.9\% on benchmarks. Embodied Reasoning~\cite{durante2024agentaisurveyinghorizons} fuses multimodal inputs with actuation for physical adaptation, while Social Reasoning~\cite{weng2023agent} employs graph embeddings for multi-agent coordination. These reveal fundamental trade-offs: symbolic methods ensure precision but constrain flexibility, while pure LLM approaches prioritize adaptability with hallucination risks.
 
Self-evolution architectures enable autonomous optimization of cognitive structures (e.g., workflows, logic modules) through closed-loop mechanisms, reducing human intervention while improving adaptability. Hu et al.~\cite{hu2025automateddesignagenticsystems} pioneer this with meta-agents programming novel agents in Turing-complete code space. Liu et al.~\cite{liu2024dynamicllmpoweredagentnetwork} formalize multi-agent collaboration as computational graphs with dynamic pruning. Liu et al.~\cite{10.1007/978-981-96-8186-0_1} further enhance edge LLM agents via cloud-edge collaboration, where a cloud-hosted LLM and edge agent co-optimize through a lightweight adapter.

While sharing foundational perception-feedback mechanisms, these paradigms reflect distinct evolutionary pathways: computational efficiency (World Models), cognitive integrity (Reasoning Systems), and recursive metamorphosis (Self-Evolution). Future progress requires cross-paradigm synthesis and ethical resolution as systems evolve into sociotechnical partners.

\subsection{AN Agent Reference Architecture}

\begin{figure}
    \centering
    \includegraphics[width=\linewidth]{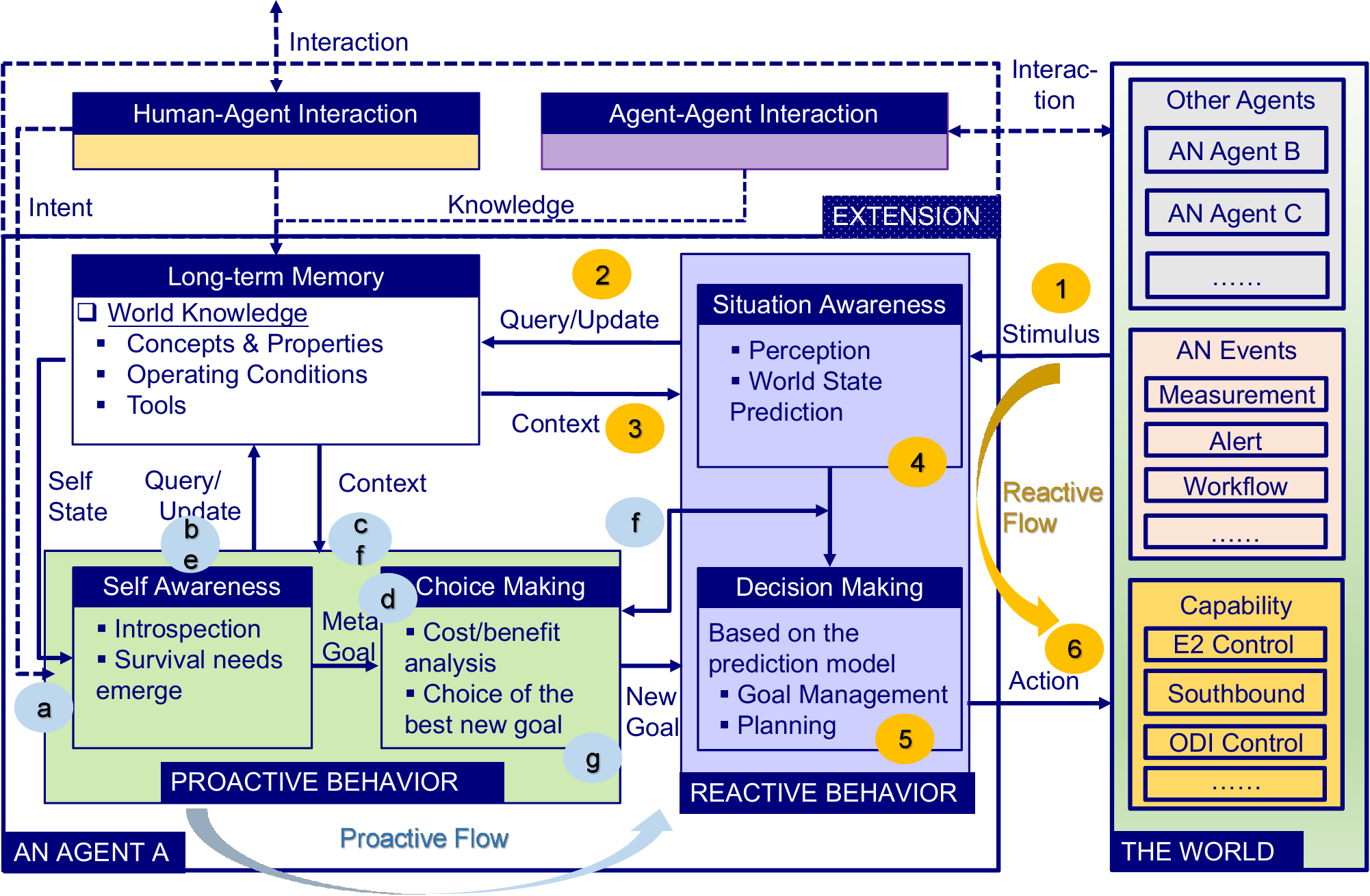}
    \caption{Sifakis's Agent architecture with proactive (processes a--g) and reactive behaviors (processes 1--6).}
    \label{fig:general_architureture}
\end{figure}

\secondRevision{
Joseph Sifakis and collaborators have established a reference architecture~\cite{sifakis2025referencearchitectureautonomousnetworks} (shown in Fig.~\ref{fig:general_architureture}) for autonomous agents that intrinsically supports both proactive and reactive behavioral modalities through integrated cognitive mechanisms. This framework adheres to a fundamental systems engineering paradigm by generalizing existing agent solutions—characterizing autonomy through the composition of mutually independent, mathematically defined cognitive functions, independent of implementation specifics. The architecture’s behavioral completeness captures the essential aspects of human cognition, particularly Kahneman’s dual-process theory~\cite{Walter2014Kahneman}, while maintaining sufficient generality to model an agent as an entity continuously interacting with its internal state and external environment (collectively the ``world''). Here, behavioral competence emerges from two interconnected systems coordinated around a central Long-Term Memory repository: the reactive subsystem generates rapid, Kahneman’s System 1-like responses to environmental perturbations, while the proactive subsystem engages in System 2-style deliberative analysis of internal state variations.}

Reactive functionality, extensively documented in domains such as autonomous robotics, operates through sensory input interpretation semantically enriched by contextual knowledge from long-term memory. For instance, when an autonomous vehicle's perception system identifies a stop sign through visual sensors, this raw sensory input activates associated procedural knowledge (``deceleration imperative'') not intrinsically embedded in the pixel data, demonstrating how contextual augmentation transforms environmental sensing into actionable intelligence. The proactive system architecturally integrates self-awareness and choice-making subsystems that operate in concert. The self-awareness module continuously monitors the agent's operational state against predefined existential objectives—such as maintaining service availability thresholds in telecommunications contexts—triggering corrective intentionality when deviations exceed tolerance boundaries. These intentions are then mapped to corresponding meta-goals by the self-awareness module. The choice-making module receives these meta-goals and selects between possible goal alternatives through cost-benefit analysis of goals satisfying the identified need.

For operational contexts requiring human supervision or multi-agent collaboration, the architecture incorporates specialized interface extensions. Human-Agent Interaction modules process operator directives and contextual network data through structured dialogue protocols, while Agent-Agent Interaction modules implement standardized coordination mechanisms for collective problem-solving. These extensions significantly enhance the framework's applicability in complex sociotechnical environments where negotiation and cooperative task execution are essential for mission success.

\section{Building Autonomous Network Agents: Implementation Framework and Technical Realization }
\subsection{Executable Runtime Instantiation }

The architecture of autonomous agents requires a clear separation between functional modules and runtime orchestration. Modules serve as deterministic input-output units—such as perception processors transforming sensory data or planners generating action sequences. Runtime systems, however, act as meta-coordination layers that dynamically sequence modules via state-machine logic. This decoupling allows flexible reconfiguration of the same modules for different domains. As the agent's cognitive core, runtime systems must interpret heterogeneous inputs (e.g., human directives or environmental telemetry), maintain contextual awareness of available capabilities (including function invocation and knowledge retrieval), and strategically navigate multidimensional solution spaces. These spaces are conceptualized as topological structures linking problem states to resolution pathways, with execution plans representing optimized trajectories through this landscape.

\begin{figure}
    \centering
    \includegraphics[width=0.9\linewidth]{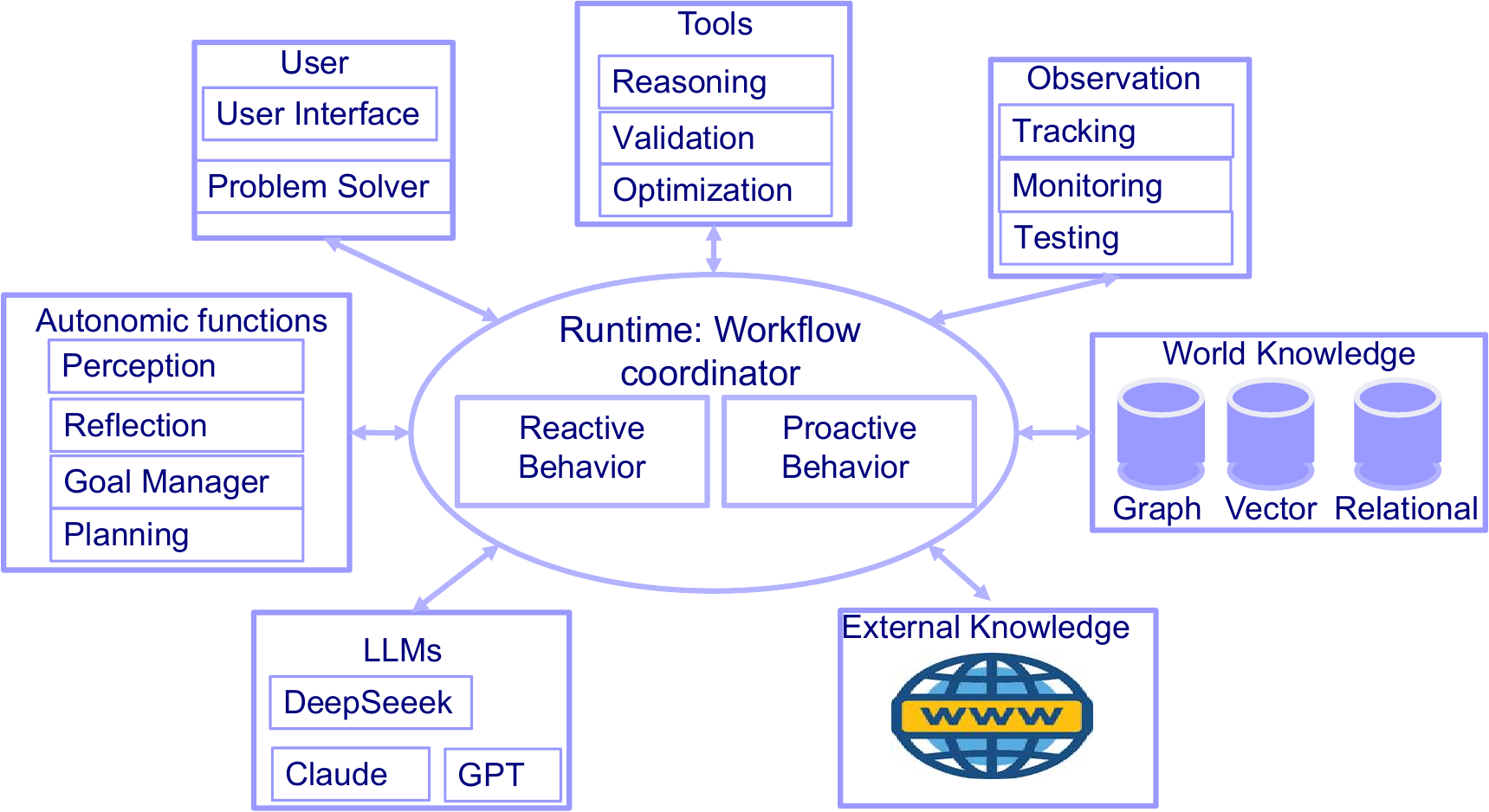}
    \caption{A flat view of an agent.}
    \label{fig:flat_view}
\end{figure}

At the architectural apex resides the Workflow Coordinator Runtime in our implementation for AN, functioning as the supreme coordination authority that dynamically allocates computational resources and governs activation sequences across specialized subsystems. This coordinator maintains a holistic state representation while managing (1) human interfaces for operator interaction, (2) analytical tools including optimization validators and formal verification engines, (3) observation pipelines processing real-time network sensory information, (4) world knowledge systems, (5) external knowledge bases, and domain-specific repositories, and (6) LLMs. This architectural nexus establishes the functional topology shown in Fig.~\ref{fig:flat_view}, where the coordinator occupies the central hub position of subsystems.

The Workflow Coordinator Runtime can invoke two runtimes: Reactive Behavior Runtime and Proactive Behavior Runtime. The Reactive Behavior Runtime (shown in Fig.~\ref{fig:general_architureture} processes 1--6) executes time-sensitive adaptations through a structured sensory-processing pipeline. When environmental stimuli are detected, the perception module generates enriched contextual interpretations that update long-term memory. This triggers iterative knowledge retrieval cycles that progressively refine contextual understanding until sufficient information is gathered to construct a predictive model. Using this model, the system generates candidate goals that undergo multi-criteria evaluation. The optimal goal selection then activates constraint-based planners to synthesize executable actions. Proposed actions are validated by the predictive model, forming a closed-loop system. This feedback mechanism continuously adjusts goals based on projected outcomes before finalizing execution plans. 

The Proactive Behavior Runtime (shown in Fig.~\ref{fig:general_architureture} processes a--g) initiates operations through continuous self-monitoring mechanisms that compare current states against predefined operational targets. When deviations exceed critical thresholds, intent generation modules synthesize operational needs by combining intrinsic objectives with externally retrieved constraints from long-term memory. These needs materialize as meta-goals (higher-order objectives), which undergo feasibility assessment against system constraints. The meta-goal is then translated into technical objectives, further justified by feasibility constraints, and the optimal goal is selected via cost-benefit analysis. The chosen goal is forwarded for management and planning, leveraging the same predictive modeling and planning infrastructure used by the reactive runtime. This architectural consistency ensures unified decision semantics across operational modes while supporting complex goal hierarchies.

\subsection{Technology Selection for Functional Modules}
{\color{black}To balance cognitive complexity with real-time responsiveness, we adopt a tiered technology stack for the functional modules:

\textbf{Hybrid Knowledge Representation:} The Long-Term Memory employs a dual-storage strategy combining graph databases (e.g., Neo4j) for structured 3GPP standards and vector databases (e.g., FAISS) for unstructured operational embeddings. This hybrid approach, augmented by Retrieval-Augmented Generation (RAG) and symbolic reasoning, ensures both logical consistency with protocols and adaptive scalability to dynamic environments.

\textbf{Hierarchical Perception and Awareness:} The Situation Awareness module utilizes a pipeline of signal conditioning techniques (e.g., Kalman filtering) to denoise telemetry, followed by LSTM networks for trend prediction. Concurrently, the Self-Awareness module leverages Large Language Models (LLMs) with few-shot prompting to translate high-level triggers—such as human directives or anomalies—into abstract meta-goals, decoupling intent understanding from implementation details.

\textbf{Decision and Control Engine:} For Choice Making, a lightweight MLP performs initial candidate ranking, while Deep Reinforcement Learning (DRL) optimizes goal selection by balancing multi-objective trade-offs via learned value functions. Finally, the Decision Making module integrates rule engines to enforce deterministic protocol safety and employs Monte Carlo Tree Search (MCTS) for optimal action planning within safety constraints.}

\section{Case Study: Radio Access Network (RAN) Link Adaptation (LA) Agent }
The performance optimization of Radio Access Networks (RAN) constitutes a fundamental challenge in achieving extreme performance metrics (e.g., ultra-reliable low-latency communication (URLLC) and enhanced mobile broadband (eMBB)) in 5G/6G networks. RAN Link Adaptation (LA) technology, due to its sensitivity to dynamic air interface conditions and millisecond-level real-time decision requirements, serves as an ideal validation case for the AN Agent architecture. \firstRevision{LA's primary mechanism involves dynamically selecting the Modulation and Coding Scheme (MCS) - a fundamental control knob that dictates both the throughput (TPT) and reliability on the wireless link. In 5G and future networks, LA must dynamically adjust MCS in complex interference environments (e.g., nonlinear power amplifier effects, inter-cell interference), diverse service requirements (TPT, reliability, latency), and multi-dimensional parameter optimization scenarios to maximize spectral efficiency while ensuring reliability.} However, traditional standardized approaches relying on post-facto feedback adjustment mechanisms are too slow to address these challenges. This case study introduces the LA Agent to validate its capability for sub-10 ms real-time closed-loop control at the network element level. By deploying intelligent Agents with local decision-making capabilities, the system continuously monitors key RAN indicators, autonomously analyzes potential air interface bottlenecks, and adjusts LA strategies accordingly. 

\subsection{Problem Modeling}
This case study focuses on the MCS optimization problem in the RAN architecture: selecting the optimal MCS for each transport block (TB) in unknown time-varying channels to maximize TPT while satisfying Block Error Rate (BLER) target constraints. The problem is inherently a real-time sequential decision-making challenge in high-dimensional stochastic environments—for each TB transmission, the base station must predict the BLER induced by candidate MCS under current channel conditions based on limited and delayed channel state information (e.g., Channel Quality Indicator (CQI) derived from User Equipment (UE) measurements). The system must then control BLER to meet both the target threshold and TPT maximization objectives.

Traditional solutions employ the Outer Loop Link Adaptation (OLLA) mechanism: predefined Signal-to-Interference-plus-Noise Ratio (SINR)-BLER mapping tables based on ideal channel models, with dynamic adjustment of SINR offsets via ACK/NACK feedback. However, this approach suffers from three critical limitations: (1) OLLA's offset adjustment relies on extensive historical feedback (typically requiring dozens of TB cycles), leading to significant convergence delays in low-traffic or high-mobility scenarios, causing prolonged suboptimal MCS selection; (2) predefined SINR-BLER models cannot accurately adapt to real-world nonlinear impairments (e.g., power amplifier distortion, phase noise, dynamic interference), particularly in millimeter-wave multi-beam scheduling scenarios, where model mismatch results in BLER control deviations; and (3) uniform offset strategies fail to accommodate differentiated service requirements (e.g., coexisting eMBB and URLLC services in a single cell requiring BLER targets of $10^{-1}$ and $10^{-3}$, respectively).

\subsection{Engineering Details}
\rebuttal{
To address the dual challenges of millisecond-level real-time control and cognitive intent understanding, our design maps the reference architecture into three distinct execution flows, enabling the system to operate efficiently on standard edge hardware while leveraging cloud-based intelligence:}
{\color{black}
\begin{itemize}
    \item \textbf{Reactive Flow (Safety Reflex):} A hard-coded safety mechanism that bypasses neural networks to enforce protocol constraints immediately. \\
    \textit{Flow: Sensor Ingest $\rightarrow$ Kalman-filter (Situation Awareness) $\rightarrow$ Rule Engine (Decision Making) $\rightarrow$ Action Command.}

    \item \textbf{Short Proactive Flow (Real-time Control):} The primary predictive control for sub-10 ms real-time inference, deployed as an xApp on the Near-RT RIC. It performs predictive optimization without changing the high-level intent. \\
    \textit{Flow: Sensor Ingest $\rightarrow$ Kalman-filter and LSTM (Situation Awareness) $\rightarrow$ RAG (Long-Term Memory) $\rightarrow$ DQN (Choice Making) $\rightarrow$ Rule Engine (Decision Making) $\rightarrow$ Action Command.}

    \item \textbf{Full Proactive Flow (Intent Changing):} An asynchronous, user-driven process deployed on the Non-RT RIC. It utilizes LLMs to translate human intent or anomalies into new policy. \\
    \textit{Flow: User input $\rightarrow$ LLM (Self-Awareness) $\rightarrow$ DQN Reward Weights Update $\rightarrow$ Short Proactive Flow.}
\end{itemize}
}

{\color{black}
The specific engineering implementation of each module is detailed below:

\noindent \textbf{Long-Term Memory Module:} This module serves as the knowledge repository, employing a dual-storage strategy. Structured 3GPP domain knowledge (e.g., MCS spectral efficiency tables) is stored in a graph database (Neo4j), while unstructured historical experiences are stored as vector embeddings in a FAISS index. To assist real-time decision-making, we implemented a Retrieval-Augmented Generation (RAG) mechanism. For every transmission, a lightweight MLP encoder encodes the current channel state, retrieving the top-5 most similar historical scenarios. These ``nearest neighbor'' experiences provide a reference distribution to guide the Choice Making module. 

\noindent \textbf{Situation Awareness Module:} This module handles the perception and prediction. Raw telemetry data first undergoes signal conditioning (sliding-window averaging for ACK/NACK smoothing and Kalman filtering for SINR). To enable proactive control, we deployed a 2-layer Bi-directional LSTM network with 128 hidden units. The model takes a sliding window of the past 100 Transmission Time Intervals and predicts the BLER trend for the next 5 TTIs. This look-ahead capability allows the agent to preemptively adjust MCS before channel degradation causes packet loss.

\noindent \textbf{Self-Awareness Module:} This module governs the agent's strategic direction. It utilizes a Large Language Model (LLM) utilizing Few-Shot Prompting to interpret natural language instructions (e.g., ``Prioritize reliability for URLLC'') into reward weights on BLER, TPT, and Latency. Here is the prompt template:

\noindent\fbox{%
    \begin{minipage}{\dimexpr\linewidth-2\fboxsep-2\fboxrule\relax}
        \scriptsize
        \textbf{System Instruction:} You are a reward function tuning expert. Generate reward weight coefficients for the DQN agent based on user instructions or system events.
        
        \textbf{Context:}
        \begin{itemize}[leftmargin=1em, nosep]
            \item Current Mode: \{mode\}
            \item Valid Ranges: $w_{\text{BLER}}, w_{\text{TPT}} \in [0,1], w_{\text{Lat}} \in [0,0.5]$
        \end{itemize}
        
        \textbf{Task:} Output ONLY a JSON dict with keys: \texttt{weight\_BLER}, \texttt{weight\_TPT}, \texttt{weight\_Latency}. Sum must be 1.0.
        
        \textbf{Few-Shot Examples (Condensed):}\\
        1. \textit{In:} ``Prioritize throughput, BLER $<$ 10\% ok'' $\rightarrow$ \textit{Out:} \texttt{\{"w\_BLER":0.3, "w\_TPT":0.7, "w\_Lat":0.0\}}\\
        2. \textit{In:} ``Extreme reliability, BLER $<$ 0.1\%'' $\rightarrow$ \textit{Out:} \texttt{\{"w\_BLER":0.8, "w\_TPT":0.1, "w\_Lat":0.1\}}
    \end{minipage}%
}

% \noindent \textbf{Choice Making Module:} As the core decision-making engine in the Short Proactive Loop, this module utilizes a {Dueling Quantile Regression DQN (Dueling QR-DQN)} architecture to robustly handle stochastic wireless channel rewards. It selects an MCS index every TTI based on a {61-dimensional state space}: real-time observations (6 dims: SINR, CQI, BLER, TPT, MCS, Latency), short-term history (48 dims: last 8 observation series), LSTM predictions (5 dims), and goal embeddings (2 dims: eMBB/URLLC one-hot). The {action space} consists of 28 discrete levels corresponding to 3GPP NR MCS indices 0--27. To prevent unsafe exploration, we implemented a RAG-guided action masking mechanism that restricts the DQN's valid actions to a maximum of the RAG-recommended MCS $+ 2$. This constraint ensures the agent operates near the historically safe region while permitting controlled optimization.{Hyperparameters} are tuned for stability: learning rate decays from $10^{-3}$ to $10^{-5}$ (Cosine Annealing); discount factor $\gamma=0.99$; a Replay Buffer (50k capacity) uses Prioritized Experience Replay (PER) to focus on critical error events; and batch size is 64. Exploration ($\epsilon$) linearly decays from 1.0 to 0.01 over 100k steps, with soft target network updates ($\tau = 0.005$) applied at every step.
\noindent \textbf{Choice Making Module:} Utilizing a Dueling QR-DQN architecture, this module selects MCS indices (0--27) based on a 61-dimensional state comprising real-time metrics (SINR, CQI, BLER, TPT, MCS, Latency), 8-step history, LSTM predictions, and goal embeddings. To prevent unsafe exploration, RAG-guided masking restricts valid actions to a maximum of the RAG-recommended MCS $+ 2$.
\textit{Configuration:} Learning rate (Cosine Annealing $10^{-3} \to 10^{-5}$); Discount $\gamma=0.99$; Batch size 64; Prioritized Experience Replay (Buffer 50k); Exploration $\epsilon$ ($1.0 \to 0.01$ over 100k steps); Soft update $\tau = 0.005$.
}

\textbf{Decision Making Module:} The module enforces protocol compliance through rule-based priority constraints (e.g., mandatory MCS reduction when BLER $> 0.1\%$), while directly executing DQN-selected MCS commands to base stations without intermediate planning.

{\color{black}
The proposed architecture explicitly fulfills the TM Forum's criteria for Level 4 ``High Autonomy,'' operating autonomously within the RAN domain while shifting the human role from manual execution to intent-based supervision. Specifically, the agent achieves \textbf{Self-Configuration} via the Self-Awareness module, which utilizes LLMs to automatically translate high-level operator intents (e.g., ``Prioritize reliability'') into mathematical reward weights, enabling zero-touch provisioning. \textbf{Self-Optimization} is realized by the Choice Making module, executing a continuous sub-10 ms closed control loop via Dueling QR-DQN to maximize spectral efficiency in real-time. Furthermore, the system demonstrates \textbf{Self-Healing} capabilities through the Situation Awareness module, where LSTM-based forecasting predicts channel degradation to proactively prevent fault events (packet loss) before they occur.
}

% \begin{table}[t]
% \caption{Mapping of Proposed Agent Capabilities to TM Forum L4 Autonomous Network Criteria}
% \label{tab:L4_mapping}
% \centering
% \begin{tabular}{|p{0.17\linewidth}|p{0.2\linewidth}|p{0.5\linewidth}|}
% \hline
% \textbf{L4 Criterion} & \textbf{Requirement} & \textbf{Our Agent's Implementation} \\ \hline
% \textbf{Scope} & Service-driven, specific domains & \textbf{Specific RAN Domain:} Autonomous Link Adaptation for eMBB and URLLC slices. \\ \hline
% \textbf{Self-Configuration} & Intent-based configuration & \textbf{Intent Translation:} LLM maps high-level intents directly to reward weights, enabling zero-touch setup. \\ \hline
% \textbf{Self-Optimization} & Closed-loop optimization & \textbf{DQN Agent:} Dueling QR-DQN executes a sub-10 ms ``Short Proactive Loop'' for real-time MCS optimization. \\ \hline
% \textbf{Self-Healing} & Predictive \& Preventive & \textbf{Predictive Control:} LSTM predicts BLER trends to proactively back-off MCS \textit{before} packet loss occurs. \\ \hline
% \textbf{Human Role} & Oversight \& Policy setting & \textbf{Intent Supervision:} Operators define goals and safety boundaries; manual parameter tuning is eliminated. \\ \hline
% \end{tabular}
% \end{table}

\subsection{Experiment Environment and Results}

\begin{figure}[!t]
    \centering
    \subfloat[Control computer, Remote Radio Units (RRUs), and Test UEs.]{
        \includegraphics[width=0.6\linewidth]{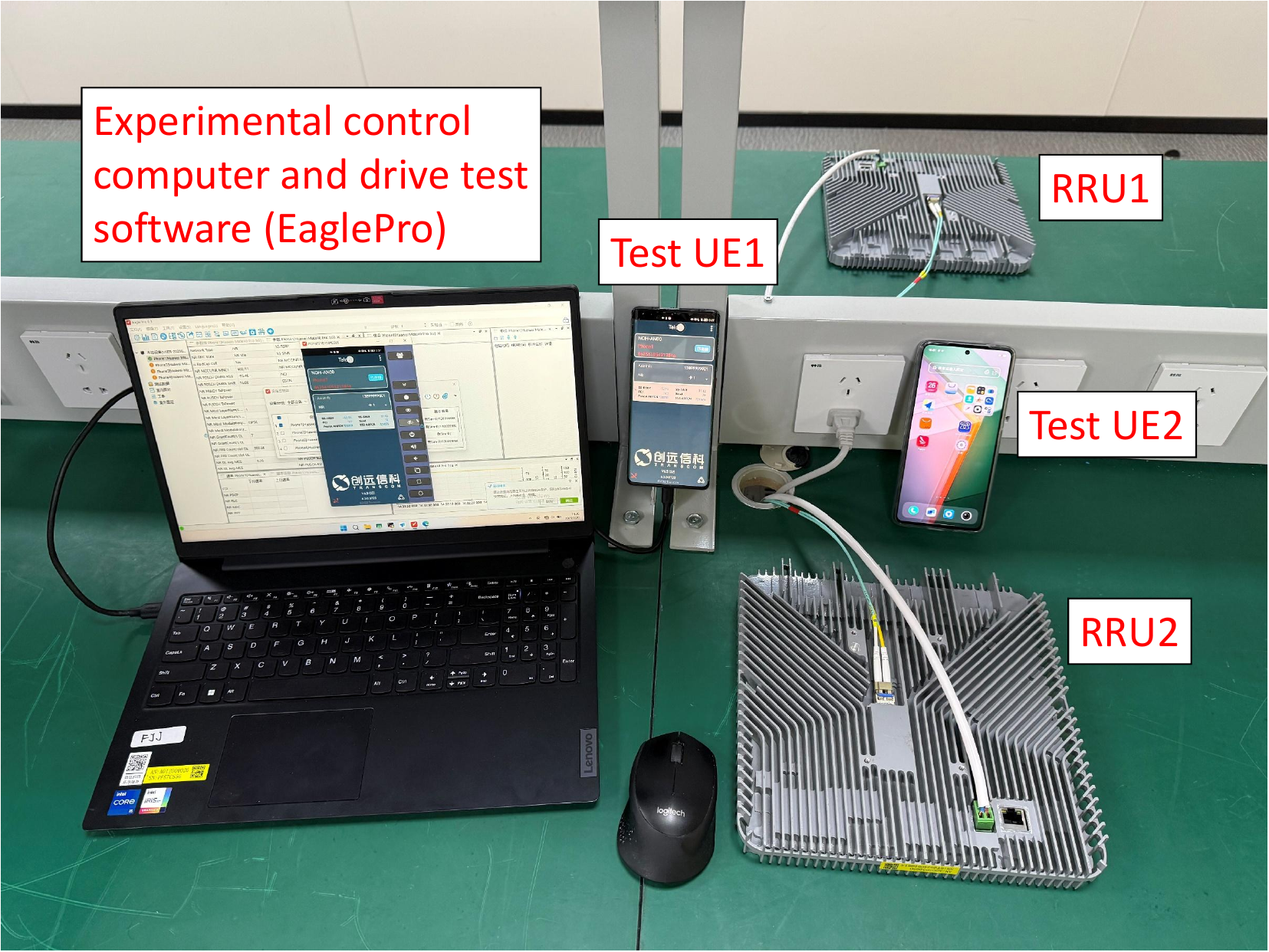}
        \label{subfig:rru_ue}
    }
    \hfill
    \subfloat[Baseband Unit (BBU).]{
        \includegraphics[width=0.6\linewidth]{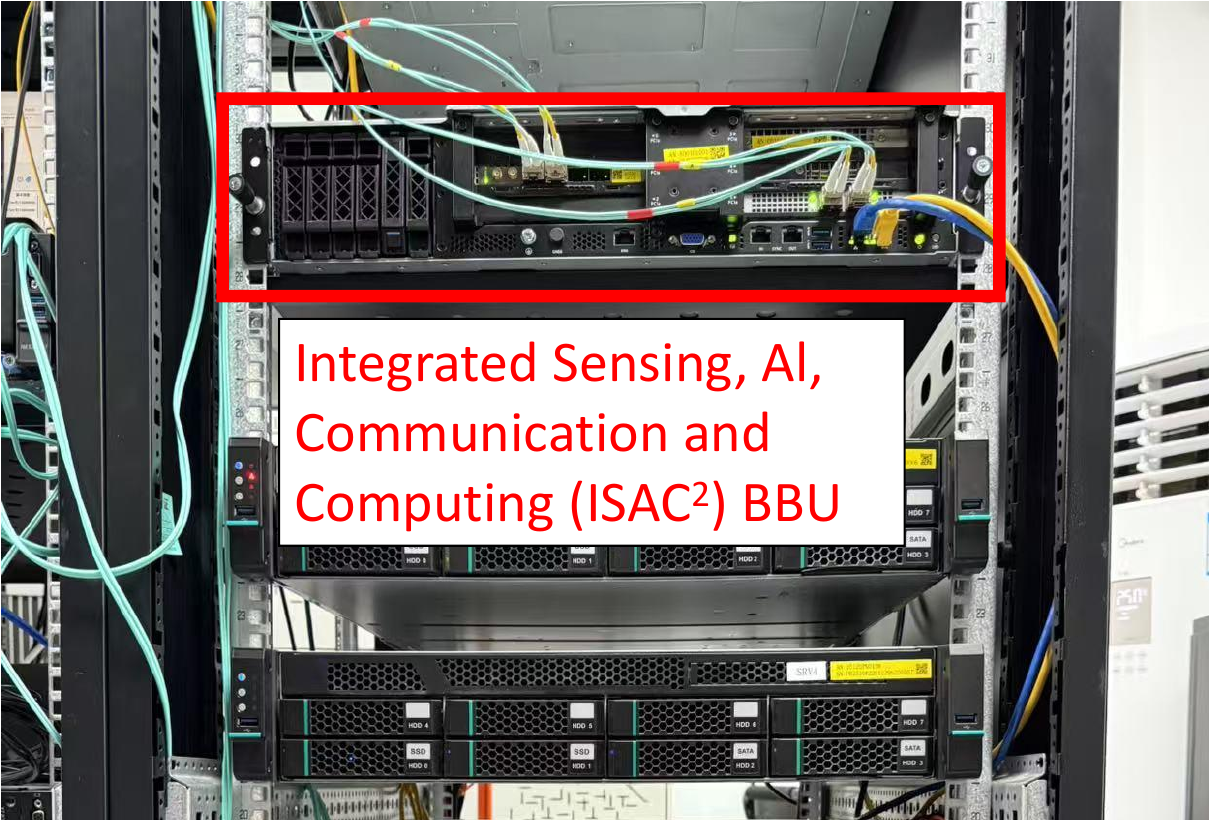}
        \label{subfig:bbu}
    }
    \caption{Test site environment. Lab wireless setup includes: 1x BBU, 2x RRUs, 2x test UEs, and 1x measurement/control host.}
    \label{fig:test_site}
\end{figure}

The experimental testbed configuration is illustrated in Fig.~\ref{fig:test_site}, featuring a control workstation running EaglePro 6.4 software for remote terminal management and comprehensive data acquisition of signaling traces and performance metrics. Two commercial 5G smartphones with distinct chipset architectures were deployed as test terminals to evaluate cross-platform compatibility: a Huawei Mate 40 Pro with software version 2.0.0.990 and a VIVO IQOO 11 running software version PD2243B\_A\_13.0.11.11.W10.V000L1. The radio access network architecture incorporated two Remote Radio Units (RRUs) connected via fiber-optic fronthaul to an ISAC$^2$ Baseband Unit (BBU), which was equipped with an Intel(R) Xeon(R) Silver 4416+ CPU operating at 2.0 GHz, 128 GB of system memory, and an NVIDIA L20 GPU with 48 GB of dedicated video memory.

\begin{figure}
    \centering
    \includegraphics[width=0.8\linewidth]{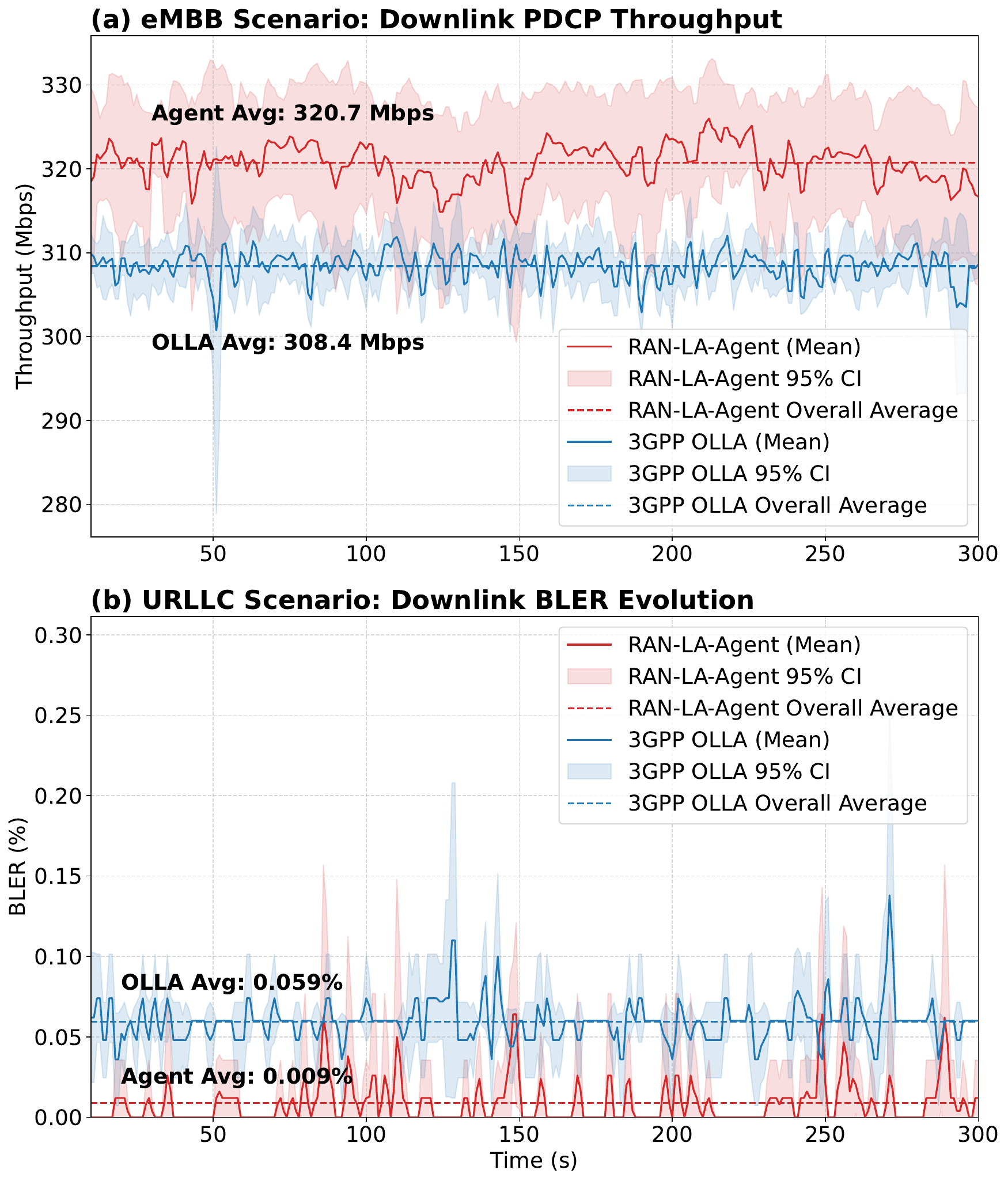}
    \caption{Comparative performance evaluation of LA Agent versus OLLA algorithm.}
    \label{fig:result}
\end{figure}

{\color{black}
Based on this hardware configuration, we verified the deployment feasibility of the proposed decoupled architecture. The \textit{Short Proactive Flow} (comprising LSTM, RAG, and DQN) functions as a lightweight xApp on the BBU. Performance metrics confirm an inference latency of 1.2--2.8 ms, comfortably fitting within the 10 ms scheduling interval of 5G frames. Resource consumption is minimal, utilizing approximately 600 MB of VRAM (only 4--5\% of the L20's capacity) and 45--95\% usage of a single CPU core for signal conditioning. Crucially, the \textit{Full Proactive Flow} hosts the computationally intensive LLM in the cloud (Non-RT RIC) as an rApp. Since intent translation is event-driven and asynchronous, it operates on a separate timescale, ensuring that large model inference never blocks the edge scheduling loop. 

The experiments were conducted in a controlled laboratory environment. This open-air testbed operated in the 3GPP N78 band centered at 3.5 GHz with 100 MHz of allocated bandwidth, maintaining controlled channel conditions characterized by Reference Signal Received Power (RSRP) levels of approximately -88 dBm with ±1 dB variation, Reference Signal Received Quality (RSRQ) measurements around -10 dB with ±0.5 dB fluctuation, and SINR values near 27 dB within a ±1 dB tolerance window. We benchmarked the proposed agent against an industry-standard OLLA baseline configured with a 100-TB sliding window, initialized via CQI feedback under a Round Robin scheduler with a maximum of 3 Hybrid Automatic Repeat Request (HARQ) retransmissions.

Data acquisition was performed using drive-test software (EaglePro), sampling Key Performance Indicators (KPIs) at high-resolution 100 ms intervals over a continuous 5-minute duration per experiment. The primary metrics included downlink Packet Data Convergence Protocol (PDCP) Throughput (Mbps) and Physical Downlink Shared Channel (PDSCH) BLER. To address the stochastic nature of wireless channels and ensure statistical significance, we conducted five independent experimental runs for each scenario. Fig.~\ref{fig:result}(a) shows the mean throughput across runs, with shaded regions representing the 95\% confidence intervals (CI). Fig.~\ref{fig:result}(b) illustrates the mean BLER along with specific error events (instances where instantaneous BLER exceeded $0.1\%$) recorded during each run. To improve visual clarity without compromising data integrity, the graphical plots employ per-second averaging.
}

{\color{black}
Fig.~\ref{fig:result}(a) analyzes the temporal throughput dynamics in the eMBB scenario. By plotting the mean performance across five independent runs with 95\% confidence intervals (shaded), we demonstrate a statistically significant advantage, evidenced by the clearly non-overlapping intervals between the LA Agent (Red) and the OLLA baseline (Blue). The Agent achieves a consistent 4\% throughput gain overall. This margin is significant given the experimental context: in a stable laboratory environment with high-end commercial test terminals, the commercial-grade OLLA baseline already delivers near-optimal throughput performance. Squeezing out additional capacity from such a saturated baseline validates the agent's superior spectral efficiency. This edge stems from the agent's sub-10 ms perception-action cycle, which enables near-instantaneous capitalization on transient spectral opportunities that the feedback-delayed OLLA mechanism inevitably misses.

In the URLLC scenario, Fig.~\ref{fig:result}(b) depicts the temporal evolution of the Downlink Block Error Rate (BLER). Unlike the eMBB scenario where the goal is capacity maximization, URLLC demands strict stability against channel stochasticity. The visualization reveals a stark contrast in robustness: the OLLA baseline (Blue) exhibits characteristic reactive instability, with frequent BLER spikes fluctuating significantly above the reliability target. Conversely, the Agent (Red) maintains a remarkably flat profile with tight confidence intervals, effectively suppressing these error peaks. Quantitatively, the Agent reduces the average BLER from 0.059\% (OLLA) to 0.009\%—an approximate 85\% reduction. This visual evidence confirms that while the baseline struggles with reactive oscillations during channel jitter, the Agent's policy successfully transitions to a proactive risk-mitigation strategy, ensuring the ultra-reliable delivery required by the slice.

To quantify the specific contributions of the cognitive modules, we conducted an ablation study isolating the Situation Awareness (LSTM) and Long-Term Memory (RAG) components. These experiments utilized the same channel conditions as the primary evaluation, measuring average performance over five independent 300-second runs. The results are summarized in Table~\ref{tab:ablation}.

\begin{table}[t]
    \centering
    \caption{Ablation Study Results (Mean Values)}
    \label{tab:ablation}
    \resizebox{\columnwidth}{!}{%
    \begin{tabular}{@{}lcc@{}}
    \toprule
    \textbf{Method / Variant} & \textbf{eMBB TPT (Mbps)} & \textbf{URLLC BLER (\%)} \\ \midrule
    {Baseline (OLLA)} & 308.4 & 0.059 \\
    \textbf{LA Agent (Full)} & \textbf{320.7} & \textbf{0.009} \\ \midrule
    \textit{w/o LSTM (No Prediction)} & 305.0 & 0.081 \\
    \textit{w/o RAG (No Knowledge)} & 311.5 & 0.012 \\ \bottomrule
    \end{tabular}%
    }
\end{table}

The analysis reveals two critical insights. First, the \textit{LSTM module constitutes the foundational capability} of the system. Removing predictive foresight (\textit{w/o LSTM}) causes performance to drop below the OLLA baseline in both throughput (305.0 vs. 308.4 Mbps) and reliability (0.081\% vs. 0.059\% BLER). This indicates that without look-ahead perception, the DQN degenerates into a reactive policy that struggles to compensate for channel feedback delays, validating that ``proactiveness" is the primary driver of gain over OLLA.

Second, the \textit{RAG module acts as a critical optimization accelerator}. While the agent without long-term memory (\textit{w/o RAG}) still outperforms OLLA, it fails to match the full agent's peak performance (311.5 vs. 320.7 Mbps). The absence of RAG-retrieved historical analogs removes the safety guardrail for the action space, forcing the agent to explore a wider, potentially riskier MCS range. This results in suboptimal convergence and a slight regression in reliability, confirming that integrating world knowledge is essential for achieving Level 4 autonomy constraints.
}

\section{Conclusion and Future Direction}
{\color{black}
This work bridges the gap between architectural theory and operational reality by implementing an Autonomous Network (AN) Agent via hybrid knowledge representation and coordinated runtimes. By integrating Dueling QR-DQN for real-time control, LSTM-based prediction, and RAG-enhanced retrieval, our system achieves a 4\% throughput gain and an 85\% reduction in block error rates over OLLA benchmarks, strictly adhering to sub-10 ms latency. This validates that cognitive capabilities can be realized in carrier-grade 6G networks without compromising protocol compliance. Future extensions will focus on timing-constrained Monte Carlo Tree Search (MCTS), ontology-guided LLM reasoning, multimodal sensing, and OWL-based interference modeling.

Looking beyond isolated optimization, the proposed architecture offers a viable path for evolving from single-agent autonomy to a collaborative {``Society of Agents.''} In this expanded scope, the {Long-Term Memory module} transcends its current role as a rule repository to become the engine for a {Knowledge-Driven Negotiation Mechanism}. By leveraging shared telecom ontologies within the graph database, agents can bridge the semantic gap between operational silos—automatically mapping abstract intents from a Core Network agent into mathematically precise constraints understood by the RAN agent. Furthermore, the causal relationships stored within the World Knowledge will enable {Coordinator Agents} to dynamically decompose high-level business intents into distributed behavior trees, facilitating the self-organization required to achieve the ``Three-Zero'' objectives in complex, multi-domain 6G environments.
}

\bibliographystyle{IEEEtran}
\bibliography{reference}

\section{Biography}

\vspace{-1cm}
\begin{IEEEbiographynophoto}{Binghan Wu}
is a Postdoctoral Researcher at AsiaInfo Technologies. He received the Ph.D. and M.Eng. degrees from the University of Sydney. His research interests include big data, knowledge representation, and artificial intelligence in autonomous networks.
\end{IEEEbiographynophoto}

\vspace{-1cm}
\begin{IEEEbiographynophoto}{Shoufeng Wang}
is Head of the Technology Innovation Center at AsiaInfo Technologies. He received the Ph.D. and B.S. degrees from Beijing University of Posts and Telecommunications. He actively contributes to standardization in 3GPP, ITU, ETSI, O-RAN, and IEEE, serving as Editor or Rapporteur in multiple projects and Chair of IEEE workshops.
\end{IEEEbiographynophoto}

\vspace{-1cm}
\begin{IEEEbiographynophoto}{Yunxin Liu}
is a Guoqiang Professor at Tsinghua Institute for AI Industry Research (AIR) and Director of the Tsinghua-AsiaInfo Joint Research Center. He received the Ph.D. degree from Shanghai Jiao Tong University and the M.S. degree from Tsinghua University. Previously, he served as Principal Researcher at Microsoft Research Asia. He serves as a general co-chair of MobiHoc, an Associate Editor of TMC, and a Vice Chair of ACM China Council SIGBED Chapter. He is an IEEE Fellow.
\end{IEEEbiographynophoto}

\vspace{-1cm}
\begin{IEEEbiographynophoto}{Ya-Qin Zhang}
is Chair Professor of AI Science and Dean of the Institute for AI Industry Research (AIR) at Tsinghua University. He received the Ph.D. degree from the George Washington University. He served as President of Baidu and held senior leadership positions at Microsoft. He is an IEEE Fellow and an elected member of the Chinese Academy of Engineering (Foreign), the American Academy of Arts and Sciences, and the Australian Academy of Technology and Engineering (Foreign).
\end{IEEEbiographynophoto}

\vspace{-1cm}
\begin{IEEEbiographynophoto}{Joseph Sifakis}
is Emeritus Senior CNRS Researcher at Verimag. He received the Ph.D. degree from the University of Grenoble. He co-received the 2007 Turing Award for contributions to model checking in system design. He has been elected to the French Academy of Sciences, French National Academy of Engineering, the Academia Europaea, the American Academy of Arts and Sciences, the American National Academy of Engineering, and the Chinese Academy of Sciences (Foreign).
\end{IEEEbiographynophoto}

\vspace{-1cm}
\begin{IEEEbiographynophoto}{Ye Ouyang}
is Chief Technology Officer and Senior Vice President at AsiaInfo Technologies. He received the Ph.D. degree from Stevens Institute of Technology and M.S. degrees from Tufts University and Columbia University. He is a Distinguished Visiting Professor at Tsinghua University, serves on the President’s Leadership Council at Stevens Institute of Technology, and he is an IEEE Fellow.
\end{IEEEbiographynophoto}

\end{document}